\documentclass{bmvc2k}

\title{Perception Visualization: Seeing Through the Eyes of a DNN}

\addauthor{Loris Giulivi \\ Mark James Carman \\ Giacomo Boracchi}{}{1}

\addinstitution{
    Politecnico Di Milano
}

\runninghead{Giulivi, Carman, Boracchi}{Perception Visualization}

\def\eg{\emph{e.g}\bmvaOneDot}

\usepackage{tabularx}
\usepackage{amsmath}
\usepackage{graphicx}
\usepackage{wrapfig}
\usepackage{amssymb}
\usepackage{array}
\usepackage{makecell}
\usepackage{hyperref}
\usepackage{xcolor}
\definecolor{giacomo}{RGB}{244,67,54}
\definecolor{loris}{RGB}{0,150,136}
\definecolor{mark}{RGB}{236,64,122}
\definecolor{todo}{RGB}{213,0,249}
\newcolumntype{?}{!{\vrule width 1pt}}
\newcolumntype{Y}{>{\centering\arraybackslash}X}

\newcommand{\mo}[1]{\textcolor{loris}{REMOVED}}

\begin{document}

\maketitle

\begin{abstract}
Artificial intelligence (AI) systems power the world we live in. Deep neural networks (DNNs) are able to solve tasks in an ever-expanding landscape of scenarios, but our eagerness to apply these powerful models leads us to focus on their performance and deprioritises our ability to understand them. Current research in the field of explainable AI tries to bridge this gap by developing various perturbation or gradient-based explanation techniques. For images, these techniques fail to fully capture and convey the semantic information needed to elucidate why the model makes the predictions it does.
In this work, we develop a new form of explanation that is radically different in nature from current explanation methods, such as Grad-CAM. \emph{Perception visualization} provides a visual representation of what the DNN perceives in the input image by depicting what visual patterns the latent representation corresponds to. Visualizations are obtained through a reconstruction model that inverts the encoded features, such that the parameters and predictions of the original models are not modified. Results of our user study demonstrate that humans can better understand and predict the system's decisions when perception visualizations are available, thus easing the debugging and deployment of deep models as trusted systems.

\end{abstract}

\begin{figure}[t]
    \centering
    \includegraphics[width=0.8\textwidth]{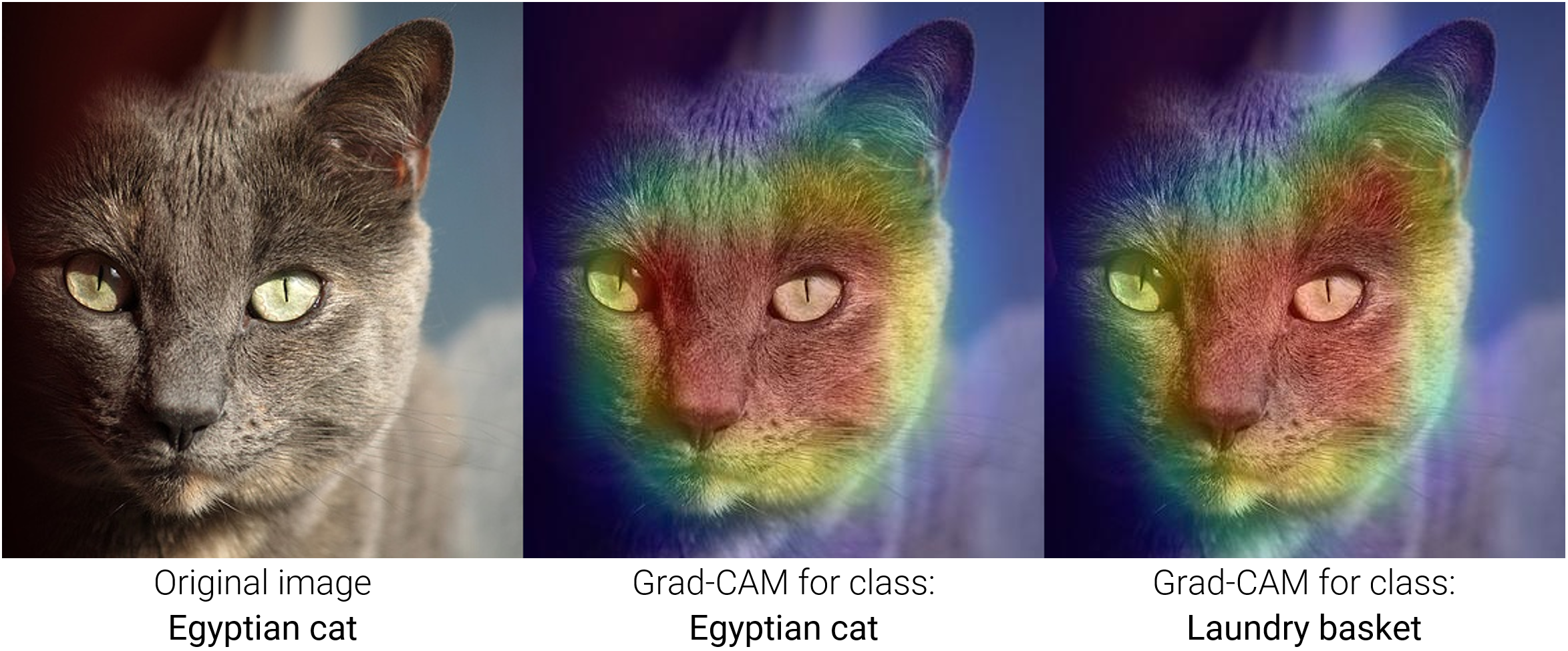} 
    \caption{Based on saliency maps it is unclear why this image is labelled as a \emph{cat} rather than a \emph{laundry basket}. Grad-CAM \cite{selvaraju2017grad} explanations are essentially the same for both classes.} 
    \label{fig:catlaundry}
\end{figure}

\section{Introduction}
Explaining a deep model is an intricate problem that requires balance between soundness and completeness to be effective in real-world scenarios such as model debugging \cite{kulesza2013too}, and is essential for building trusted systems. Explanations need to succinctly convey the model's reasoning and not overwhelm the user. Current explanation techniques such as Grad-CAM \cite{selvaraju2017grad} have focused on generating pixel-wise measures of conspicuity, by analysing the effect each pixel has on the model's prediction using gradient or perturbation-based analysis \cite{das2020opportunities}. However, these are often uninformative~\cite{rudin2019stop}, as exemplified in Figure \ref{fig:catlaundry}. We believe that explanations should instead carry semantic meaning at a higher level. Thus, we introduce \emph{Perception Visualization} (PV), a novel technique to explain the latent semantics of a deep convolutional neural network (CNN). 

\begin{figure}[t]
    \centering
    \includegraphics[width=0.96\textwidth]{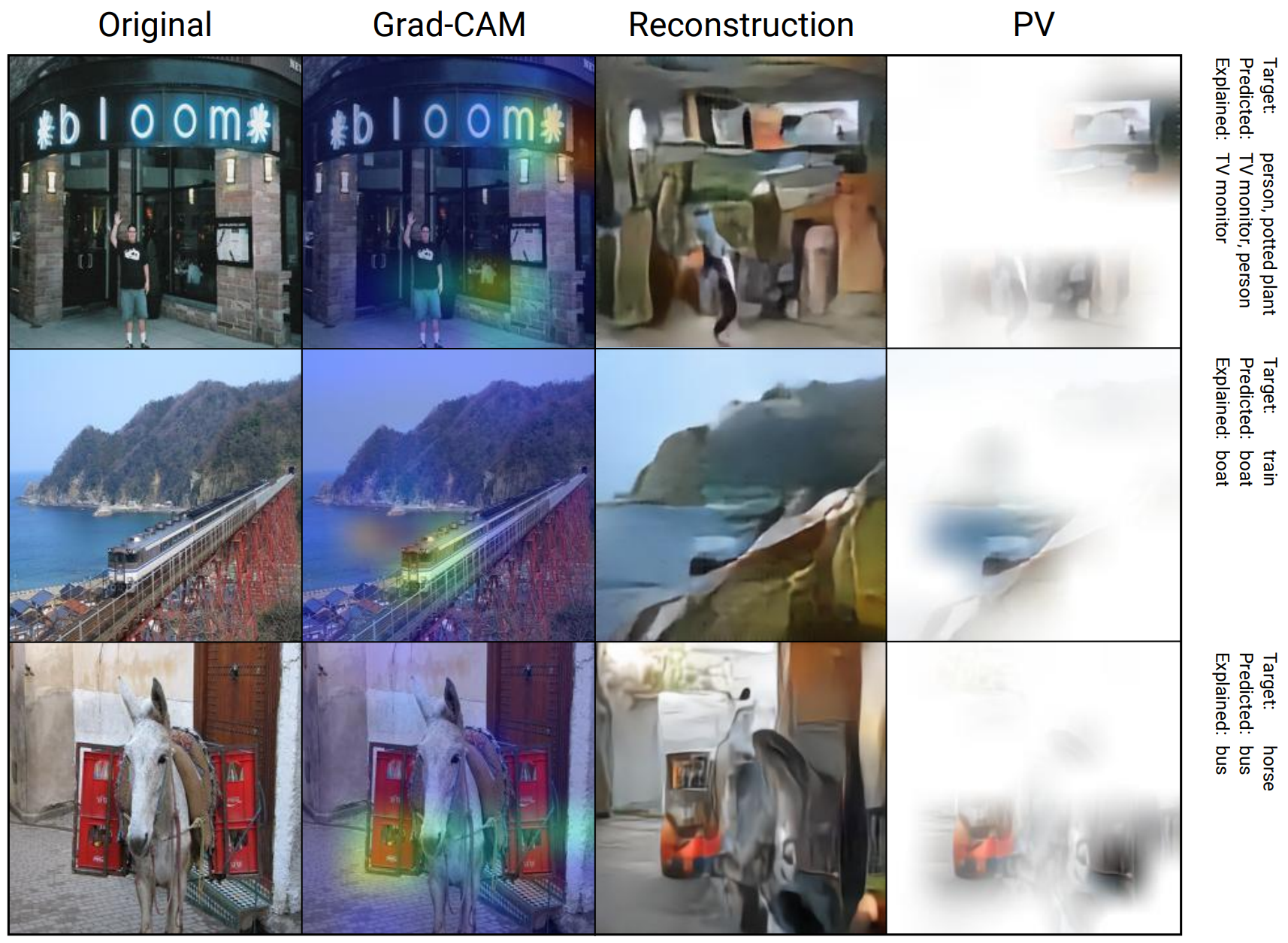} 
    \caption{Misclassified examples and their explanations using Grad-CAM and PV. Explanations given by PV are much more informative and are able to describe the network's error by depicting semantic features of the network's perception. In particular, the three PVs depict: a rectangular object recalling a TV screen, a black and white object in water recalling a boat, and an orange shape recalling a bus.} 
    \label{fig:misclassified-examples}
\end{figure}

PV consists of two components: \emph{i}) a gradient-based saliency map and \emph{ii}) a reconstruction obtained through network inversion. This combination allows PV to show both \emph{where the model is looking} and \emph{what the model is seeing}, in contrast to the vast majority of previous techniques that only show \emph{where} the network is focusing its attention when making a decision. To the best of our knowledge, ours is the first work producing image explanations by using a neural network to invert latent representations. Moreover, our work aims at providing explanations for the diagnostic situation in which a data scientist performs error analysis on an image classifier, manually inspecting images on which the model performs poorly. \emph{Useful explanations should inform the direction for resolving the fault}, such as procuring more training data from a particular domain. In Figure~\ref{fig:misclassified-examples} we see examples of such misclassified images, and note that the PV for the first image shows that the model is confusing a neon sign for a television, which indicates a possible lack of training images containing neon signs. Such a realization could \emph{not} have been obtained without the help of the perception visualization, since neon signs aren't even a class in this problem. In this case, we see that PV explanations can be employed in a \emph{prescriptive manner} in order to improve model performance.

PV reveals itself empirically to be particularly effective at explaining incorrect predictions from the model, as shown in Figure \ref{fig:misclassified-examples}. 
We validate PV through a user study, investigating the users' ability to guess the model's predictions when explanations are given. Results of a survey on circa 100 subjects show that PV is able to help respondents better determine the predicted class in cases where the model had made an error.
We make our code publicly available at: \url{https://github.com/loris2222/PerceptionVisualization}

\section{Related Work}
\label{sec:related}
Explainable artificial intelligence (XAI) is an emerging field that is experiencing a surge in research interest. We now discuss XAI techniques as they relate to image classification.

An explanation, in the context of XAI, is a way to present results to a human in understandable terms \cite{doshi2017towards}. This definition is purposely vague and includes, for example: textual descriptions of the reasoning behind the prediction \cite{kim2018textual}, heatmaps indicating the pixels that most contributed to the result \cite{selvaraju2017grad}, or graphs that match decisions with some knowledge base \cite{wang2021shapley}. Our explanations are local to a single sample, similarly to what is done in Grad-CAM \cite{selvaraju2017grad} and SHAP \cite{lundberg2017unified}, and differently to works such as LIME \cite{ribeiro2016should} and SpRAy \cite{lapuschkin2019unmasking}. Moreover, PV differs from methodologies that make use of gradient information (\eg Grad-CAM \cite{selvaraju2017grad}) or perturbation analysis (\eg RISE \cite{petsiuk2018rise}, Score-CAM \cite{wang2020score}), in that we use a deep neural network to reconstruct the semantics of the latent space.

The most prominent techniques to explain images in XAI literature are saliency maps, which attempt to visualize \emph{where} the model places its attention in the image. These techniques include class-activation maps~\cite{zhou2016learning,oquab2015object} and their subsequent improvements~\cite{selvaraju2017grad,morbidelli2020augmented,chattopadhay2018grad}. A saliency map is itself an image, usually of the same size as the input, highlighting where the model focuses its attention. Given a saliency map, an explanation can be constructed either by superimposition or by masking. In the former case, the saliency map is displayed with different hues depending on the importance of the region. For the latter, the image is covered and only regions deemed relevant by the saliency map are shown. A more detailed description of these practices is provided in the supplementary material.

Other works, which are closer to ours in philosophy, have attempted to directly visualize \emph{what the model sees} in the input image rather than simply \emph{where it focuses its attention}. Google's \textit{Inceptionism} \cite{mordvintsev2015inceptionism} and Simonyan et. al. \cite{simonyan2013deep} optimize an input image to maximize the network's response to some desired class, and in turn give an intuition of how the model represents such class. Our work, instead, provides visualizations of the model's perception for each image sample. Perhaps the most relevant, HOGgles \cite{vondrick2013hoggles} inverts visual features, most notably allowing to view images through \textit{HOG glasses (HOGgles)}. This work, however, relies on feature dictionaries to solve an optimization problem to reconstruct small image windows. Our work, on the other hand, uses a neural network to decode full images, only requires optimization during the learning phase, and works on much more complex models.

\section{Perception Visualization}
Our core contribution is \emph{perception visualization} (PV): a novel explainability technique that makes use of a deep neural network to explain a pre-trained deep neural network. PV visually displays the input of the model \emph{from the model's point of view}, which is that of its latent representation. PV leverages CNN inversion techniques and displays the resulting reconstructions in a way that can help users understand whether the model correctly perceived the input or not. 
Several works, including \cite{mahendran2016visualizing,dosovitskiy2016generating,zhao2016loss}, propose techniques for reconstructing the original image from its latent representation. These take ideas from research in autoencoders \cite{bank2020autoencoders} and generative adversarial networks (GAN) \cite{goodfellow2014generative} to improve reconstruction quality. In our work, we show how network inversion can be used to generate novel explanations, providing a better insight into the model's functioning without modifying its performance. 

\begin{figure}[t]
    \centering
    \includegraphics[width=0.9\textwidth]{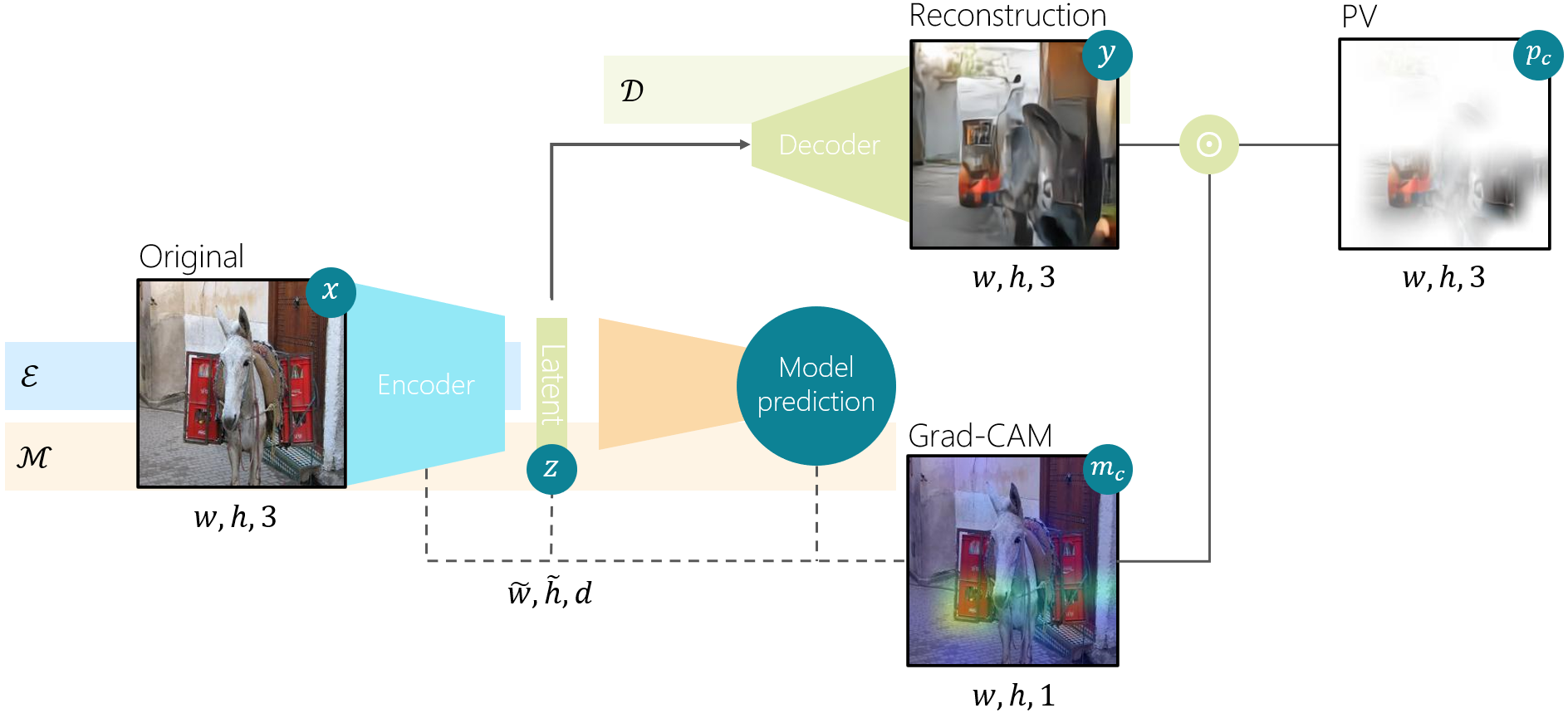} 
    \caption{An overview of our method and interactions between the models involved. Encoder $\mathcal{E}$ is a truncation of the model $\mathcal{M}$ which we want to explain, decoder $\mathcal{D}$ is trained to reconstruct the encoder's latent representations. From these, we compute Grad-CAM saliency maps and reconstructions, which are then combined to obtain PV.}
    \label{decoder architecture}
\end{figure}

\vspace{0.2cm}
\noindent\textbf{Method:}
In what follows we describe Perception Visualization for CNN classifiers (Figure \ref{decoder architecture}), which are the most widely used deep neural models for images. Let us denote the trained CNN we want to explain as:
\begin{equation}
  \mathcal{M}: \mathbb{R}^{w,h,3}\, \rightarrow \mathbb{R}^n, 
\end{equation}
which maps RGB images of size $(w,h)$ into a vector of $n$ posterior probabilities. PV is a class-discriminative explanation that we define, for a class $c \in \{ 1,\dots, n\}$ and for an input image $x\in \mathbb{R}^{w,h,3}$, as a second image $p_c \in \mathbb{R}^{w,h,3}$ constructed such that each of its pixels $p_c(i,j,k)$ in position $(i,j)$ of the $k$-th color channel is defined as:
\begin{equation}\label{eq:pv}
p_c(i,j,k) \doteq (1-m_c(i,j)) + m_c(i,j) y(i,j,k)\,,
\end{equation}
where $m_c \in \mathbb[0,1]^{w,h,1}$ is a saliency map for class $c$ and $y \in \mathbb{R}^{w,h,3}$ is the reconstruction obtained through network inversion. In the remainder of this section, we will describe these two components in detail.

\vspace{0.2cm}
\noindent\textit{\textbf{Saliency Map:}}
The saliency map $m_c \in \mathbb[0,1]^{w,h,1}$ is computed from $x$ with respect to a specific class $c$. While any saliency map algorithm could in principle be used, we adopt Grad-CAM \cite{selvaraju2017grad} in our experiments. The map depicts \emph{where} the model is looking, namely which portions of $x$ have influenced the prediction for class $c$. The saliency map is used to determine which regions of the reconstruction to show in the PV. By definition \eqref{eq:pv}, all the pixels where $m_c(i,j) = 0$ are mapped to white pixels in the explanation, while pixels where $m_c(i,j) = 1$ return the corresponding values in the reconstruction, namely $y(i,j,k)$. 

\vspace{0.2cm}
\noindent\textit{\textbf{Reconstruction:}}
The reconstruction component $y \in \mathbb{R}^{w,h,3}$ of PV is responsible for displaying \emph{what} the model is seeing, and it is obtained by training a network to invert the feature extraction portion of $\mathcal{M}$. Starting from $\mathcal{M}$, we define its submodel $\mathcal{E}$, namely the \emph{encoder}:
\begin{equation}\label{eq:encoder}
 \mathcal{E}: \mathbb{R}^{w,h,3} \rightarrow \mathbb{R}^{\widetilde{w},\widetilde{h},d} \text{,}
\end{equation}
which, given an image $x$, computes its latent representation $z = \mathcal{E}(x),\, z \in \mathbb{R}^{\widetilde{w},\widetilde{h},d}$ having spatial dimensions $(\widetilde{w},\widetilde{h})$ and depth $d$. The encoder is merely a truncation of the original model which we define to ease the description of our methods. Thus, \emph{no network re-training is required}, and the latent representation $z$ computed by $\mathcal{E}$ is the same for PV and for inference.

We call the \textit{decoder} the model that computes the reconstruction, which is instead trained to restore the original image $x$ from the latent representation $z$:
\begin{equation}\label{eq:inversion}
    \mathcal{D}: \mathbb{R}^{\widetilde{w},\widetilde{h},d} \rightarrow \mathbb{R}^{w,h,3} \text{.}
\end{equation}
In this setup, given an input sample $x$, we define its reconstruction $y$ as: 
\begin{equation}
y=\mathcal{D}(\mathcal{E}(x))\,.
\end{equation}
In our experiments we have inverted a pre-trained ResNet-50~\cite{he2016deep} model, using as latent representation the output of its deepest convolutional layer ($conv5\_block3$), which is a tensor of spatial dimensions $[7,7]$ and depth 2048. Our decoder is therefore a CNN made of blocks of convolutional layers and transposed convolutions to perform image up-sampling. The kernel size was set to $[3, 3]$ for all layers, and all the layers have leaky ReLU activation with $\alpha=0.2$, except for the output layer which uses a sigmoid activation and the transposed convolutional layers which are linearly activated. We also perform batch normalization after each transposed convolution. An overview of our model architecture and further discussion regarding our design choices are detailed in the supplementary material.

Lastly, we note that while we have defined PV for CNN-based classifiers, our choice was only guided by the popularity of this kind of models as a benchmark for explainability techniques. Indeed, PV is extensible to any architecture that allows latent space reconstruction and saliency map computation. This includes, but is not limited to, CNN-based models for captioning and segmentation. We discuss possible extensions of our work in Section \ref{sec:futurework}.

\vspace{0.2cm}
\noindent\textbf{Decoder Training:}
We train the decoder to invert latent representations generated by pre-trained models. The decoder, both during training and inference, has only access to the embeddings $\mathcal{E}$, which renders reconstructions dependent on the model $\mathcal{M}$ to be explained.
Inspired by recent studies \cite{dosovitskiy2016generating}, we train $\mathcal{D}$ using different terms in the loss function: 
\begin{equation}\label{eq:loss}
      \mathcal{L} =  \alpha_1 \ \mathcal{L}_{MSE} \ + \ \alpha_2 \ \mathcal{L}_{SSIM} \ + \ \alpha_3 \ \mathcal{L}_{DSIM}\,,
\end{equation}
where $ \alpha_i \geq 0 \ i = 1,\dots,3$ are tuning parameters that we force to sum to one to disentangle the norm of the loss from the learning rate. The components of the loss function are defined in an attempt to yield reconstructions that are faithful to the latent representation. For the same reason, we do to allow skip connections (such as those of U-nets \cite{ronneberger2015u}), as doing so would enable the decoder to use information from layers further away from the latent representation.
The following losses are for batches of input images $X = \{x_i\}$ of size $b$, their latent representations $Z = \mathcal{E}(X)$ and their reconstructions $Y = \mathcal{D}(\mathcal{E}(X))$.

\vspace{0.2cm}
\noindent\textbf{• Reconstruction Error:} The MSE between the input images and the recovered images:
\begin{equation}
    \mathcal{L}_{MSE} \doteq \sum_{i=1..b}~\sum_{j,k,l}|y_i(j,k,l)-x_i(j,k,l)|^2 \text{.}  
\end{equation}

\noindent\textbf{• Structural similarity loss:}   SSIM \cite{wang2004image} is a full reference metric that is widely used in image restoration, and that has also been used to train CNNs \cite{zhao2016loss}. SSIM measures the similarity of images from the perspective of perceived change in structural information, and exploits measures of luminance and contrast. User studies show how SSIM correlates better than MSE with visual quality assessment \cite{wang2004image}. 
The SSIM loss is obtained by negating the sum of the SSIM index over each channel of an RGB image:
\begin{equation}
  \mathcal{L}_{SSIM} \doteq -\sum_{i=1..b}~\sum_{c\in\{R,G,B\}}SSIM_c(x_i,y_i) \text{.}   
\end{equation}

\noindent\textbf{• Deep perceptual SIMilarity loss (DSIM):}
When training a decoder using only image-space losses, it is possible that the decoder learns to invert any embedding to the input image, independently of the correctness of the prediction. The DSIM component counters this effect by forcing reconstructions and input images to be similar \emph{in latent space}. This translates in a similarity between the encoding $z = \mathcal{E}(x)$ of the input and the encoding $\mathcal{E}(y)$ of the output. Thus, following Dosovitskiy \& Brox~\cite{dosovitskiy2016generating}, 
we define the DSIM loss as the $L_2$ norm between the original latent representation and the latent representation of the reconstructed image:
\begin{equation}
    \mathcal{L}_{DSIM} \doteq  \sum_{i=1..b}\sum_{j,k,l}|\mathcal{E}(y_i)(j,k,l)-z_i(j,k,l)|^2 \,.
\end{equation}

\begin{figure}[t]
    \centering
    \includegraphics[width=0.9\textwidth]{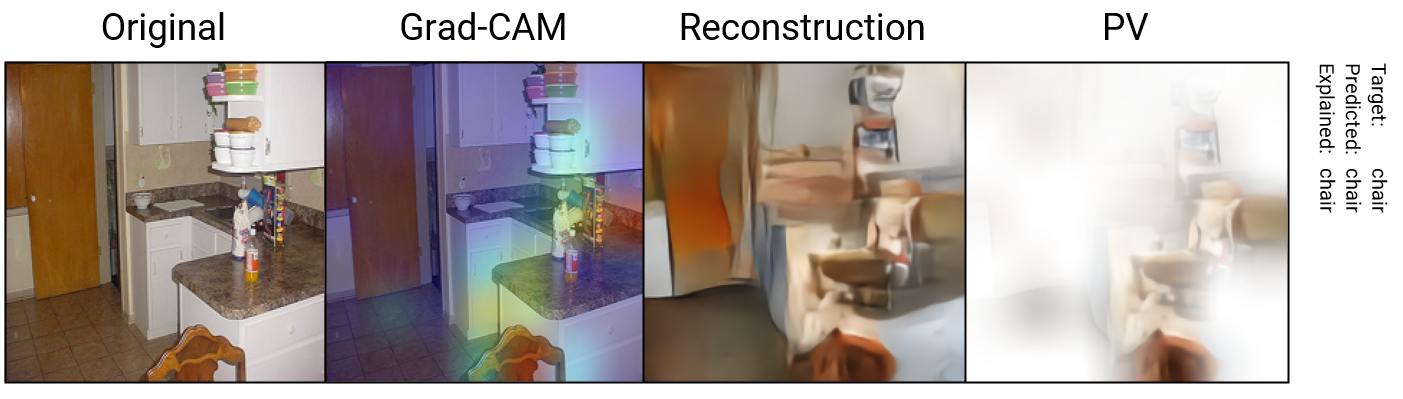}
    \caption{Grad-CAM for class ``chair'' shows that the model has correctly recognized the chair at the bottom of the image, but that also the table was part of the region of interest. PV explains how the network was fooled by reconstructing a second chair in that region.}
    \label{figure:more chairs}
\end{figure}

We have also run experiments including a Wasserstein GAN \cite{arjovsky2017wasserstein} loss term, but found that, due to the very large and sparse nature of the latent space, these were not beneficial in our case. Our best model achieves similar reconstruction quality as in~\cite{dosovitskiy2016generating} (model without GAN) using the hyper-parameter combination $\alpha_1 = 0.2, \alpha_2 = 0.4, \alpha_3 = 0.4$. We discuss hyper-parameter tuning in the supplementary material, including the topic of GAN loss terms.

\vspace{0.2cm}
\noindent\textbf{PV as an Explanation:}
The intuition behind PV is that whenever the model correctly predicts an output label, the latent features must be consistent with the input. Consequently, the reconstruction will be similar to the input. Instead, whenever the model makes a mistake, activations in the latent space are not consistent with the input, and the reconstruction will semantically resemble those erroneous features, as observed in the PVs of Figure \ref{fig:misclassified-examples}.

Another advantage of PV over techniques such as Grad-CAM is that PV can explain faults in saliency maps even when the prediction is correct, as shown in Figure \ref{figure:more chairs}. Indeed, there are cases where the localization seems to make little sense, despite the correctness of the model's prediction. Whenever this happens, PV will provide insight on the network's output. The user, looking at the reconstruction, can determine which features of the image were most difficult for the network to correctly perceive.


Lastly, we mention that, in contrast to saliency maps, PV requires to train an additional network ($\mathcal{D}$). We argue, however, that this is a small price to pay to overcome the inherent limitations of plug-and-play explanations, such as those discussed in \cite{rudin2019stop}. Moreover, we stress that training only involves $\mathcal{D}$, while \emph{$\mathcal{M}$ and its performance remain unaltered}.


\section{Experiments}
\label{sec:experiments}
\begin{figure}[t]
    \centering
    \includegraphics[width=0.7\textwidth]{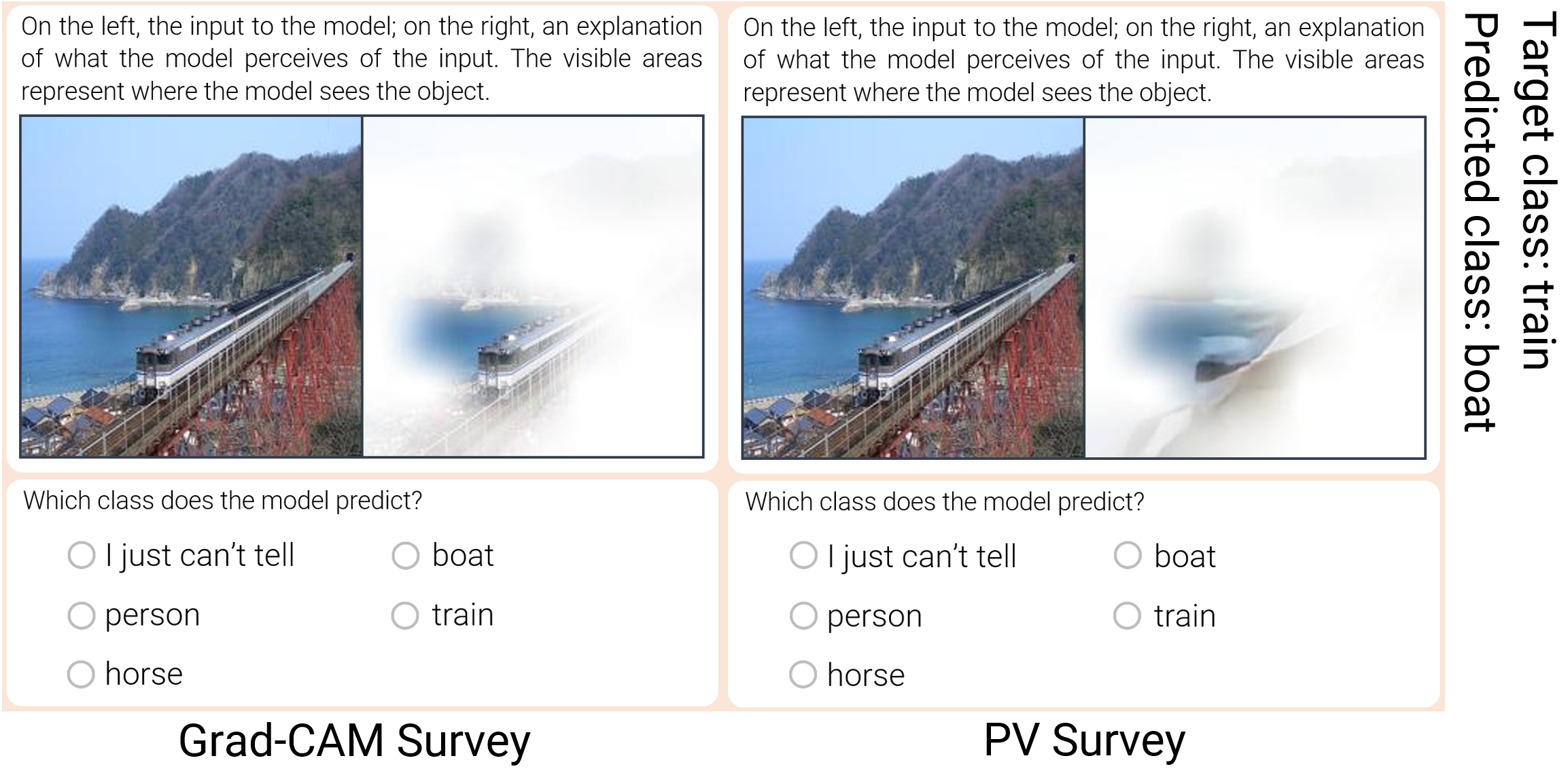}
    \caption{Example question from our surveys. Note how the possible answers are the same for PV and Grad-CAM surveys (the order is randomized in the user's view).}
    \label{survey example}
\end{figure}
The main objective of explainability techniques is to make the user understand why the model provided some output. A best practice to quantitatively assess the effectiveness of an explainable model is to measure \textit{simulatability} \cite{doshi2017towards,alqaraawi2020evaluating,hase2020evaluating}, which means measuring how often users are able to guess the model's prediction when given the explanation to be evaluated (possibly paired with the input). 
The core of our experiments comes in the form of a survey, composed of a multiple choice quiz, following in the footsteps of \cite{alqaraawi2020evaluating}. During the survey, the participants are asked, given different explanations, to predict the model's output. To compare with other state-of-the-art explanations, two different surveys have been administered to different people, one showing PV and the other showing Grad-CAM explanations. Furthermore, we perform additional experiments demonstrating the resilience of PV's semantic value to changes in the decoder.

\vspace{0.2cm}
\noindent\textbf{Survey Structure:}
The structure of the survey is identical for both PV and Grad-CAM, and is composed of a first portion introducing the explainability technique and of a series of questions. The first section is meant to help users understand what they are about to see. Then, a series of 30 input images are shown paired with their explanation, and for each of them we ask: \emph{``What is the model's prediction?''}. The user can then select one of five options, amongst which will be the class (or classes) that is actually present in the image (i.e. the true label) and the class that the model predicts, which will differ when the model misclassified the sample. The remaining options are chosen randomly. It is important to note that throughout the survey we do not inform the user whether or not the model has made a mistake. Figure \ref{survey example} shows an example question from our surveys. Survey structure, question images and respondent's answers are detailed in the supplementary material.

\vspace{0.2cm}
\noindent\textbf{Experiment Details:}
We run the experiment by applying PV and Grad-CAM on a pre-trained ResNet-50 \cite{he2016deep} model performing transfer learning on the popular PASCAL VOC \cite{pascal-voc-2012} dataset, due to its use in other similar works \cite{alqaraawi2020evaluating}, the small number of classes (needed for non-dispersive surveys), and its multi-label nature (which renders the task more complicated). 
Question images are uniformly sampled between correctly classified (16 samples), and incorrectly classified (14 samples). Due to the multi-label classification, we have confined our selection to those images for which the prediction and label set coincided (correctly classified), and those for which none of the targets were in the prediction set (incorrectly classified). During the survey, we provide explanations and require answers only for the top predicted class.
For each question, we avoid providing sets of options where the correct answer is very apparent and the others are clearly wrong. To do this, we select five candidate answers for each question, composed of: \emph{i}) the model's prediction, \emph{ii}) three other target labels for the sample (i.e. the objects actually present in the image), and \emph{iii}) a ``I just can't tell'' option. In the rare case where there were an excess of true labels, some were dropped, and likewise when there were insufficient labels, random classes were chosen (from those in the dataset). The order of answer options was randomized for each question and user. \emph{Additionally, to counteract possible response biases,  choices are the same for both surveys}.

A web application assigned users to the surveys in a round-robin fashion. Survey participants were gathered via university mailing lists sent to graduate students enrolled in machine learning related subjects. We expect participants to be somewhat literate in artificial intelligence. No personally identifying information was collected from participants.

\vspace{0.2cm}
\noindent\textbf{Survey Results:}
We gathered responses from a total of $98$ participants. The two surveys were administered in a round-robin fashion, therefore, we expected the same number of responses from the Grad-CAM and the PV surveys. However, responses were uneven: $40$ for the PV survey, and $58$ for the Grad-CAM survey. As the survey was administered on-line, we do not know how many people dropped out before the end of the survey, however, this number is higher for the PV survey, and may be due to the poor reconstruction quality. As a baseline, we expect random guessing (excluding the ``I just can't tell'' option) to score $25\%$ overall, both on the correct and incorrect subsets. Instead, if the explanations provided were to be uninformative, we expect users to gather all their information from the original image, and to predict the model's output well only when the model made a correct prediction.

\begin{table}[t]
    \caption{Aggregated participants' accuracy at determining the model's predictions on the 14 questions where the model had predicted incorrectly and the 16 questions where it predicted correctly for the two explainers: Grad-CAM and PV.}
    \centering
    \resizebox{0.80\textwidth}{!}{
        \begin{tabular}{|c|c|c|}
            \hline
             &  \multicolumn{2}{c|}{\textbf{User Accuracy ($\pm$ standard error)}}\\
             &  \multicolumn{2}{c|}{
                \begin{tabular}{cc}
                     Model's prediction is \textbf{incorrect}  & Model's prediction is \textbf{correct} \\
                \end{tabular}
            }\\
            \hline
            Grad-CAM &  ~~~~~~~~~~~~~~~8.3\% (±1.3\%)~~~~~~~~~~~~~~~  & \textbf{95.9}\% (±0.7\%) \\
            PV & \textbf{35.0}\% (±3.0\%) & 75.5\% (±2.5\%) \\
            \hline
        \end{tabular}
    }
    \label{results table}
\end{table}

\begin{figure}[t]
    \centering
    \includegraphics[width=0.80\textwidth]{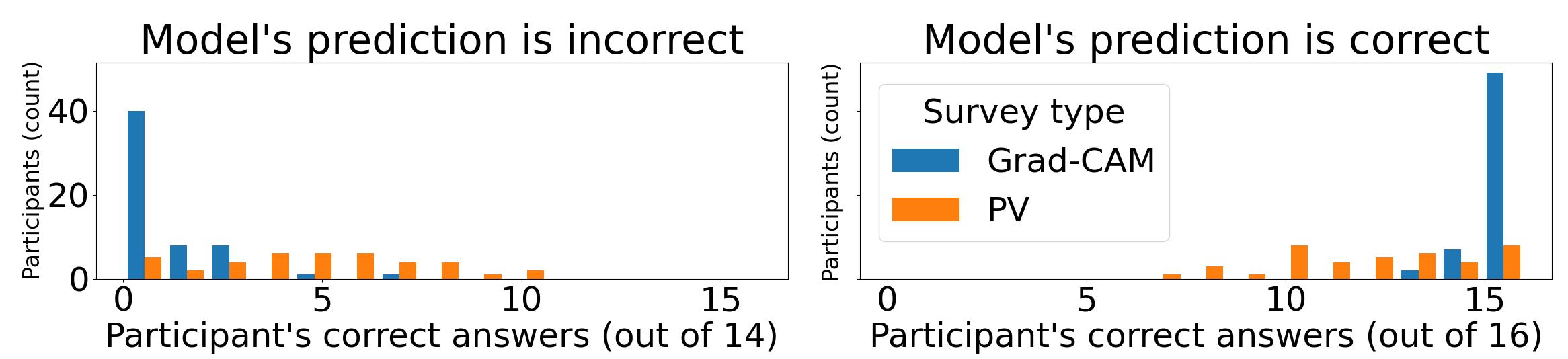} 
    \caption{Histograms of participant's accuracy in terms of the number of times they determined the models prediction for cases where the model was \emph{incorrect} in its prediction (left) and \emph{correct} in its prediction (right).
    }
    \label{figure:histogram}
\end{figure}

Our results (Table \ref{results table}, Figure \ref{figure:histogram}) show that \emph{respondents to the PV survey were better at determining which class the model was predicting when this made a mistake}, case for which performance improved from 8.3\% for Grad-CAM to 35.0\% for PV. For samples that were instead correctly classified, due to the lower visual clarity of PV, Grad-CAM users were better at determining the predicted class by 95.9\% compared to 75.5\%.

We used the Mann-Whitney U test to check if there was a statistically significant difference in accuracy between users of the PV and of the Grad-CAM versions of the survey. For cases where the model was \emph{incorrect}, the left tailed test ($H_1$: Acc(Grad-CAM)$<$Acc(PV)) confirms that the performance improvement achieved by PV is significant ($p=2.3e^{-11}$, $Z=-6.58$). For cases where the model gave a \emph{correct} prediction, the finding is in the opposite direction: a right tailed test ($H_1$: Acc(Grad-CAM)$>$Acc(PV)) suggests that, for this case, Grad-CAM was significantly better ($p=2.45e^{-11}$, $Z=6.57$).

Overall, the performance improvement seen in cases where the model is wrong is in line with the objective of providing insight for the purpose of debugging a model, that is, whenever its predictions are incorrect. For other use-cases, we recommend pairing PV and Grad-CAM explanations together to leverage their respective stregths. 
Lastly, these results suggest that the encoder and decoder are not separate entities, that is, that \emph{the decoder has not merely learned to imitate the input}. Indeed, if this were the case, users would not have been able to correctly predict the model's output when the input was misclassified.

\vspace{0.2cm}
\noindent\textbf{Invariance to decoder training:}
\begin{figure}[t]
    \centering
    \includegraphics[width=0.65\textwidth]{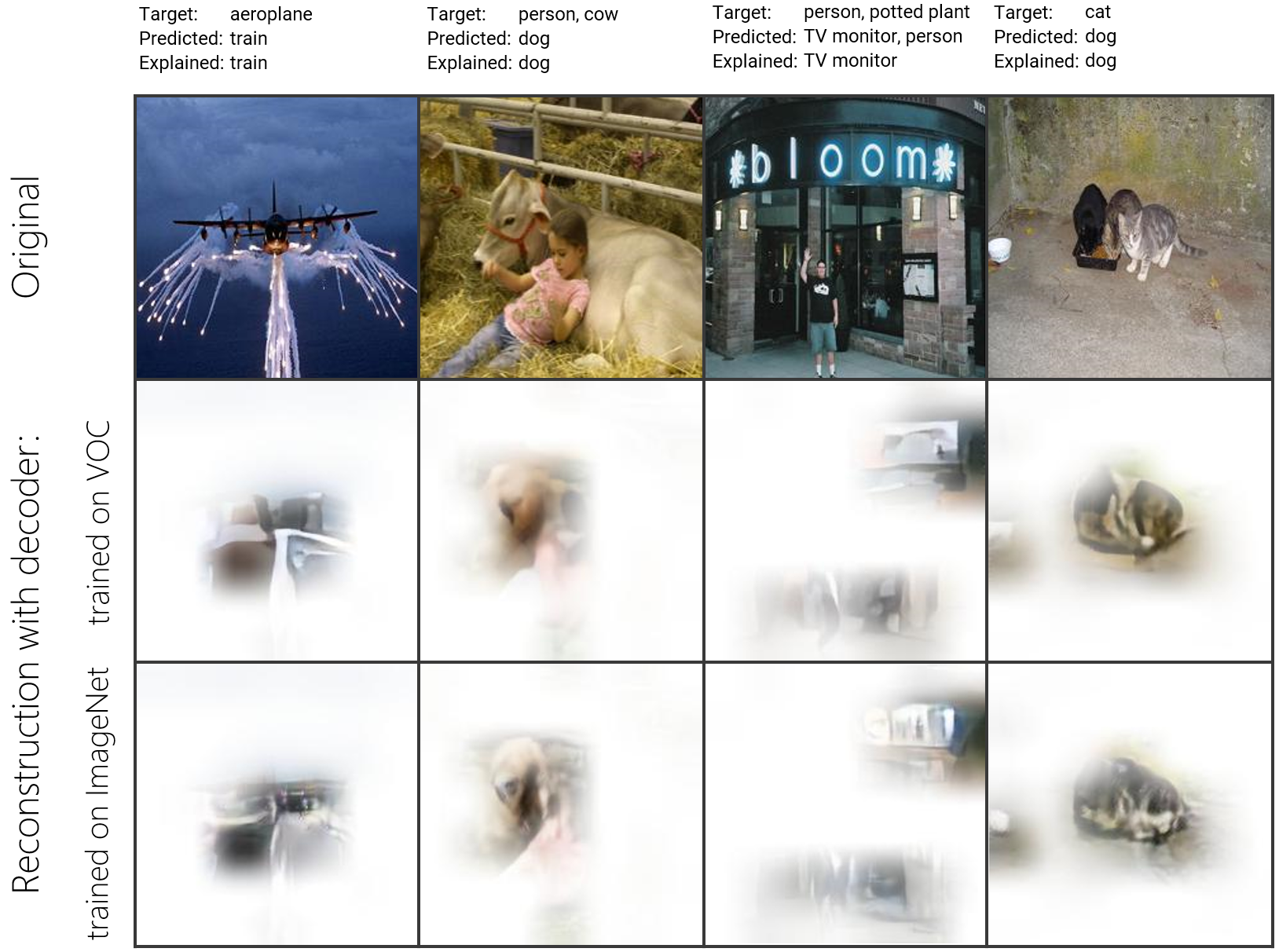} 
    \caption{Reconstructions remain semantically similar between decoders, even when the decoder is trained using a different dataset than that used to train $\mathcal{M}$.}
    \label{figure:othertrainingdataset}
\end{figure}
Since PVs depend on the result of the decoder training process, we investigate whether explanations remain consistent under different training conditions. On the one hand, this would verify that the decoder's output is strongly linked to $\mathcal{M}$'s latent representation, thus to the network's perception. On the other hand, this would ensure practical use of PV in different circumstances, such as that of using a different dataset.



The decoder $\mathcal{D}$ used in our previous experiment was trained using the same dataset used to also train the model $\mathcal{M}$ to be explained (PASCAL VOC). For this experiment, we train a new decoder by using a dataset composed of a random subset of $8\,000$ ImageNet \cite{ILSVRC15} images (a sample count very similar to the $8\,077$ VOC images used for training $\mathcal{M}$). Since we are always using a ResNet-50 model to encode images, also ImageNet samples are cropped and resized to $224\times224$.

We provide qualitative evaluation of the results in Figure \ref{figure:othertrainingdataset}, showing that reconstructions are only marginally altered, and still possess all the key features necessary for explaining the model's mistakes. In particular, we note (Figure \ref{figure:othertrainingdataset}) how features resembling the train in the first sample remain (even though in a less clear way) also when the decoder is trained on ImageNet. In the second and fourth samples, the dog features are present also in the ImageNet trained decoder, and possibly even more clearly in the second sample. Finally, in the third sample, we see how features pertaining the TV monitor are visible in both decoders.

These results suggest that the semantics of the decoder's reconstructions are preserved in two successful training runs in different conditions. This, in turn, indicates that PVs are heavily determined by $\mathcal{M}$'s latent representation, rather than the specific decoder training. Furthermore, this experiments shows that it is possible to train $\mathcal{D}$ using a different dataset than that used to train $\mathcal{M}$, and that the two datasets might not even have the same class sets (as is the case for ImageNet and PASCAL VOC).

\section{Conclusions and Future Work}
\label{sec:futurework}
We have introduced Perception Visualization (PV): the first method to provide explanations by exploiting a neural network to invert latent representations. PV provides semantically relevant information by learning to visualize the model's perception. As shown in our experiments, this allows PV to increase the user's performance in predicting the model's output in cases where the model misclassifies a sample, hence giving better insight on the model's functioning than what was previously achievable using only saliency maps.

An important future direction consists in improving reconstructions, firstly in terms of visual clarity, but also in terms of faithfulness to the latent representation. So far, we were not able to improve reconstruction quality with GANs due to the nature of the space that needs to be inverted, however, further studies regarding the properties of the latent representations may allow to overcome this problem. We also plan to investigate class-discriminative decoding to provide a broader insight on the model, and exemplar losses to generate prototype-based reconstructions \cite{chen2018looks}.
We believe that better reconstructions could enable the application of PV also in medical imaging and in other critical domains.

Other promising directions regard the extension of PV to different tasks and architectures other than classification. For example, class-discriminative decoding could be used to explain recurrent/transformer networks used for image captioning \cite{cornia2020m2}. In this context, we would generate sequences of PVs, one for each output token.

\bibliography{bibliography}
\end{document}